\documentclass{article}

\usepackage{microtype}
\usepackage{graphicx}
\usepackage{subfigure}
\usepackage{booktabs} 

\usepackage{hyperref}

\usepackage[accepted]{icml2024}

\usepackage{amsmath}
\usepackage{amssymb}
\usepackage{mathtools}
\usepackage{amsthm}

\usepackage{styles/symbols}
\usepackage{comment}
\usepackage{wrapfig}

\usepackage[capitalize,noabbrev]{cleveref}

\theoremstyle{plain}
\newtheorem{theorem}{Theorem}[section]

\theoremstyle{definition}

\theoremstyle{remark}

\usepackage[textsize=tiny]{todonotes}

\icmltitlerunning{Neural-Kernel Conditional Mean Embeddings}

\begin{document}

\twocolumn[
\icmltitle{Neural-Kernel Conditional Mean Embeddings}




\begin{icmlauthorlist}
\icmlauthor{Eiki Shimizu}{soken,ism}
\icmlauthor{Kenji Fukumizu}{ism}
\icmlauthor{Dino Sejdinovic}{adelaide}
\end{icmlauthorlist}

\icmlaffiliation{soken}{Department of Statistical Science, Graduate University of Advanced Studies (SOKENDAI), Tokyo, Japan}
\icmlaffiliation{ism}{The Institute of Statistical Mathematics, Tokyo, Japan}
\icmlaffiliation{adelaide}{School of Computer and Mathematical Sciences, The University of Adelaide, Australia}

\icmlcorrespondingauthor{Eiki Shimizu}{shimizu.eiki@ism.ac.jp}

\icmlkeywords{Kernel Methods, Neural Networks, Conditional Distribution}

\vskip 0.3in
]



\printAffiliationsAndNotice{\icmlEqualContribution} 

\begin{abstract}
Kernel conditional mean embeddings (CMEs) offer a powerful framework for representing conditional distribution, but they often face scalability and expressiveness challenges. In this work, we propose a new method that effectively combines the strengths of deep learning with CMEs in order to address these challenges. Specifically, our approach leverages the end-to-end neural network (NN) optimization framework using a kernel-based objective. This design circumvents the computationally expensive Gram matrix inversion required by current CME methods. To further enhance performance, we provide efficient strategies to optimize the remaining kernel hyperparameters. In conditional density estimation tasks, our NN-CME hybrid achieves competitive performance and often surpasses existing deep learning-based methods. Lastly, we showcase its remarkable versatility by seamlessly integrating it into reinforcement learning (RL) contexts. Building on Q-learning, our approach naturally leads to a new variant of distributional RL methods, which demonstrates consistent effectiveness across different environments.
\end{abstract}

\section{Introduction}
\label{submission}
When conditional distributions present complexities such as multimodality, skewness, or heteroscedastic noise, simply capturing the conditional mean no longer suffices. Kernel conditional mean embeddings (CMEs) \citep{Song2009CME,muandet2017kernel} represent conditional distributions as elements within a reproducing kernel Hilbert space (RKHS), a space associated to a positive definite kernel. Given both input and output variables, CMEs map these variables into a high-dimensional (often infinite) feature space using kernels, enabling nonparametric and flexible characterization of conditional distributions. Indeed, CMEs have been successfully employed in settings such as probabilistic inference tasks \citep{Fukumizu2013KBR,song2013kernelembeddings} and causal inference tasks \citep{Singh2019KIV,MuaKanSaeMar21,ChaTonGonTehSej2021,ParShaSchMua21}.

However, CMEs possess several limitations. First, Gram matrix inversion, needed for standard CME estimation, can become prohibitively expensive when the dataset grows large. Second, the pre-specified nature of RKHS features can lead to poor performance when dealing with high-dimensional variables exhibiting highly nonlinear structures. Lastly, standard hyperparameter tuning procedures such as cross-validation are not applicable for selecting hyperparameters of the kernel on output variables. This last limitation arises because the objective function is defined in terms of the RKHS norm associated to this kernel, and any change in kernel parameters fundamentally alters the objective function itself.

To address these challenges, we propose a method that effectively blends deep learning with CMEs. The central concept involves replacing computationally demanding matrix inversion with a neural network (NN) model, while simultaneously leveraging the feature learning capabilities of deep learning. Our method integrates seamlessly with standard NN training procedures, differing only in its use of an RKHS-based loss function. Additionally, we propose using a specific type of kernel on the output variable, and introduce supplementary objective functions tailored to optimize the kernel parameters. Notably, we offer two flexible options for the hyperparameter optimization: iterative or joint optimization with the NN training process.

The true potential of CMEs lies in their applicability beyond standard density estimation tasks. Namely, CME representations can be employed to define metrics over the space of conditional probability distributions \citep{Gretton2012MMD}. This powerful property, coupled with the efficient incorporation of NNs within our approach, paves the way for effortless adoption within deep reinforcement learning (RL) contexts. By modeling discounted sum of rewards using our NN-based CME representation, we naturally arrive at a new variant of distributional RL \citep{bellemare2017distributional,bdr2023} methods. Our approach shares traits with two existing methods, CDQN \citep{bellemare2017distributional} and MMDQN \citep{Nguyen-Tang_Gupta_Venkatesh_2021}, while demonstrating consistent efficacy across three benchmark environments.

The paper is organized as follows: \cref{sec:preliminary} introduces kernel mean embeddings and CMEs, including a NN-parameterized deep feature approach. \cref{sec:proposed-method} presents our NN-CME hybrid approach and two hyperparameter optimization strategies. \cref{sec:experiment} demonstrates the performance of our method on both synthetic and real-world datasets. \cref{sec:RL} delves into RL, introducing our new distributional RL method and assessing its performance in three classic control environments.

\section{Background and Preliminaries} \label{sec:preliminary}
{\bf Notations:} 
We consider random variables $X$ and $Y$, residing in domains $\mathcal{X}\subseteq\mathbb{R}^{d_x}$ and $\mathcal{Y}\subseteq\mathbb{R}^{d_y}$, respectively, with realizations $x$ and $y$, joint distribution $P$, and density function $p(x,y)$. Measurable positive definite kernels $k_\calX: \calX \times \calX \to \mathbb{R}$ and $k_{\calY}: \calY \times \calY \to \mathbb{R}$, associated with RKHSs $\calH_\calX$ and $\calH_\calY$, induce features $\psi(x) = k_\calX(x, \cdot)$ and $\phi(y) = k_\calY(y, \cdot)$. The Inner product is denoted by $\braket[\calH]{\cdot,\cdot}$ and the RKHS norm by $\|\cdot\|_{\calH}$.

\subsection{Kernel Mean Embeddings}
For a given marginal distribution $P(X)$, the kernel mean embedding (KME) $\mu_{P(X)}$ is defined as the expectation of the feature $\psi(X)$:
\begin{align*}
    \mu_{P(X)} = \expect[X]{\psi(X)} \in \calH_\calX,
\end{align*}
and always exists for bounded kernels. The reproducing property of RKHSs equips KMEs with the useful ability to estimate function expectations: $\braket[\calH_\calX]{f, \mu_{P(X)}} = \expect[X]{f(X)}$ for any $f\in\calH_\calX$. Furthermore, for \emph{characteristic kernels}, these embeddings are injective, uniquely defining the probability distribution \cite{fukumizu2008kernel,Sriperumbudur2010KME}. For instance, popular kernels such as Gaussian and Laplace kernels possess this property.

The second order mean embeddings, also known as covariance operators \cite{fukumizu2004dimensionality}, are defined as the expectation of tensor products between features, for instance: 
\begin{align*}
    \covXY \!=\! \expect[XY]{\psi(X)\!\otimes\!\phi(Y)},
\end{align*}
 where $\otimes$ is the tensor product, and always exist for bounded kernels. Covariance operators generalize the familiar notion covariance matrices to
accommodate infinite-dimensional feature spaces.

\subsection{Kernel Conditional Mean Embeddings}
With the aforementioned building blocks in place, we now extend KMEs to to the case of conditional distributions, defining kernel conditional mean embeddings (CMEs) \citep{Song2009CME,muandet2017kernel} as follows:
\begin{align*}
    \mu_{P(Y|X)}(x) = \expect[Y|x]{\phi(Y)|X=x} \in \calH_\calY,
\end{align*}
requiring an operator $C_{Y|X}:\calH_\calX \rightarrow \calH_\calY$ that satisfies: (a) $\mu_{P(Y|X)}(x) = C_{Y|X} \psi(x)$ and (b) $\braket[\calH_\calY]{g, \mu_{P(Y|X)}(x)} = \expect[Y|x]{g(Y)|X=x}$ for $g\in\calH_\calY$.
 Assuming $\expect[Y|x]{g(Y)|X=\cdot}\in \calH_\calX$, the following operator satisfies the requirements:
 \begin{align*}
    C_{Y|X} = (\covXX)^{-1}\covXY.
\end{align*}
Given i.i.d samples $\{(x_i, y_i)\}_{i=1}^n \sim P$, an empirical estimate of the operator can be obtained as follows:
\begin{align*}
    \hat{C}_{Y|X} &=  (\hatcovXX + \lambda I)^{-1}\hatcovXY\\
    &= \Phi(K_X+ \lambda I)^{-1}\Psi^\top,
\end{align*}
where $\lambda>0$ is a regularization parameter, $K_X$ is the Gram matrix $(K_X)_{ij} = k_\calX(x_i, x_j)$, and $\Psi$ and $\Phi$ are the feature matrices stacked by columns: $\Psi = [\psi(x_1), \dots,  \psi(x_n)]$ and $\Phi = [\phi(y_1), \dots,  \phi(y_n)]$. Alternatively, this empirical estimate can be obtained by solving following function-valued regression problem \cite{grunewaelder2012conditional}:
\begin{align}
    \argmin_{C:\calH_\calX \rightarrow \calH_\calY} \frac1n \sum_{i=1}^n \left\Vert\phi(y_i) - C\psi(x_i)\right\Vert_{\calH_\calY}^2 + \lambda \left\Vert C\right\Vert_{HS}^2,  \label{eq:fuction_loss}
\end{align}
where $\|C\|_{HS}$ is the Hilbert-Schmidt norm. Putting together these elements, we arrive at the empirical estimator:
\begin{align}
    \hat{\mu}_{P(Y|X)}(x) =\hat{C}\psi(x)=\sum_{i=1}^n\beta_i(x)\phi(y_i)
    = \Phi\vec{\beta}(x), \label{eq:CME_empirical}
\end{align}
where:
\begin{align*}
\vec{\beta}(x)=(K_X+ \lambda I)^{-1}\vec{k}_X,
\end{align*}
with $\vec{k}_X=[k_{\calX}(x_1,x), \dots,  k_{\calX}(x_n,x)]^\top$.
In contrast to KMEs for marginal distributions, CMEs employ non-uniform weights  $\beta_i(x)$, which are not constrained to be positive or sum up to one.

\subsection{Deep Feature Approach}
Instead of relying on pre-specified RKHS feature maps, the Deep Feature approach (DF) \citep{xu2021learning} harnesses the adaptive capabilities of deep learning to learn tailored feature representations. This approach has demonstrated its efficacy in settings such as causal inference \citep{xu2021learning} and kernelized Bayes' rule \citep{pmlr-v162-xu22a}, often outperforming classical CME methods.

By replacing $\psi$ with a $d$-dimensional NN-parameterized feature map $\psi_\theta: \calX \to \mathbb{R}^{d}$ (where $\theta$ denotes the NN's parameters) in \eqref{eq:fuction_loss}, we can jointly optimize both the feature representation and the conditional operator $C$. This is achieved by optimizing $\theta$ with the following loss function:
\begin{align*}
    \hat{\mathcal{L}}(\theta)
    = \mathrm{tr}\left(K_Y(I -  \Psi^\top_\theta(\Psi_\theta\Psi_\theta^\top + \lambda I)^{-1}\Psi_\theta)\right),
\end{align*}
where $K_Y$ is the Gram matrix with elements $(K_Y)_{ij} = k_\calY(y_i, y_j)$, and gradient-based optimization methods can be applied. Given the learned parameter $\hat{\theta} = \argmin_\theta \hat{\mathcal{L}}(\theta)$, we can express $\vec{\beta}(x)$ in \eqref{eq:CME_empirical} as follows:
\begin{align*}
    \vec{\beta}(x) =  \Psi^\top_{\hat\theta}\left(\Psi_{\hat\theta}\Psi^\top_{\hat\theta} + \lambda I \right)^{-1} \psi_{\hat\theta}(x).
\end{align*}
This DF approach offers a partial solution to scalability challenges as well. Its computational complexity of $O(nd^2+d^3)$ is typically smaller than the $O(n^3)$ complexity of classical approaches. During the training, mini-batch optimization can be be applied for further acceleration. However, the regularization parameter $\lambda$ plays a vital role in both performance and numerical stability, and its selection often necessitates computationally expensive cross-validation procedures involving multiple NN optimizations. Additionally, this approach on its own does not address the remaining hyperparameter selection issue for $k_\calY$.

\section{Neural Network Based CME}\label{sec:proposed-method}
\subsection{Model and Objective Function}
While DF approach of \citet{xu2021learning} partially addresses computational challenges of CME, it still restricts $\vec{\beta}(x)$ to a specific functional form involving matrix inversion. The core idea of our approach is to replace $\vec{\beta}(x)$ in \eqref{eq:CME_empirical} with $f(x;\theta): \calX \to \mathbb{R}^{M}$, a NN parameterized by $\theta$. We propose the CME estimator of the following form:
\begin{align*}
\hat{\mu}_{P(Y|X)}(x) = \sum_{a=1}^{M} \phi(\eta_{a})f_a(x;\theta),
\end{align*}
where $\eta_a\in\mathcal{Y}$ are $M$ location parameters. These parameters can be optimized together with the NN parameter $\theta$, or can be fixed to reduce the number of parameters. In this paper, we opt for fixing them throughout for easiness of optimization.
Analogous to \eqref{eq:fuction_loss} but without $\|C\|^2_{HS}$, $\theta$ is optimized through:
\begin{align*}
 \min_{\theta}\frac1n \sum_{i=1}^n \left\Vert\phi(y_i) - \sum_{i=1}^M\phi(\eta_{a})f_a(x_i;\theta)\right\Vert_{\calH_\calY}^2.
\end{align*}
We denote this as $\min_{\theta}\frac1n \sum_{i=1}^n \hat{\ell}(\theta)$, where $ \hat{\ell}(\theta)$ can be further expressed as:
\begin{align}
 \hat{\ell}(\theta)=-2\sum_{a}k_\calY(y_i,\eta_a)w_a+\sum_{a,b}k_\calY(\eta_a,\eta_b)w_aw_b,\label{eq:RKHS_loss}
\end{align}
with $w_a=f_a(x_i;\theta)$ and $w_b=f_b(x_i;\theta)$. The term $\sum_{i=1}^n k_\calY(y_i, y_i)$ is omitted, assuming fixed hyperparameters for $k_\calY$, for now. In contrast to DF, the whole process of calculating $\vec{\beta}(x)$ with explicit features is effectively amortized with $f(x;\theta)$. This eliminates the need for both Gram matrix inversion and regularization parameter selection.

However, selecting hyperparameters for $k_\calY$ remains a challenge due to the dependence of the objective values on the RKHS norm $\|\cdot\|_{\calH}$; they are not comparable for different hyperparameters. While downstream tasks can sometimes guide tuning (e.g., in IV regression: \citet{xu2021learning}), one typically resorts to heuristic methods like the median heuristic \cite{gretton2005kernel}. Our aim in the following sections is to construct a criterion that can serve as a supplementary objective function for optimizing the $k_\calY$ parameter, offering a more effective approach to its selection.

\subsection{Choice of Kernel $k_\calY$}
Our initial step is to employ a positive definite kernel $k$ that satisfies: $k(\cdot,\cdot)\geq0$ and $\int_{}^{}k(y,\cdot) dy=1$. Thus, the kernels we consider are both reproducing kernels and smoothing kernels for densities, two often confused, but distinct notions of kernel functions.  In the following, we adopt what we call the \emph{Gaussian density kernel} for $k_\calY$:
\begin{align*}
k_{\sigma}(y, y^\prime)=\left(\frac{1}{\sqrt{2\pi\sigma^2}}\right)^{d_y}\exp \left( -\frac{\|y - y^\prime\|^2}{2\sigma^2} \right).
\end{align*}
We will denote the associated RKHS by $\calH_{\sigma}$. This type of kernel has seen previous use in \citet{JMLR:v13:kim12b,pmlr-v89-hsu19a}, and variants corresponding to other kernels, such as Laplace and Student kernels, are also available. Because Gaussian density kernel is also a smoothing kernel, CME estimator also gives an estimator $\hat{p}(y|x)$ of the conditional probability density, obtained through the inner product  $\braket[\calH_{\sigma}]{k_{\sigma}(y,\cdot), \hat {\mu}_{P(Y|X)}(x)}$ as
\begin{align}\label{eq:cond_density}
 \hat{\mathbb E}_{Y|x}[k_{\sigma}(y,Y)|X=x] = \sum_{a=1}^{M} k_{\sigma}(y, \eta_{a}) f_a(x;\theta).
\end{align}
This form of density estimate has been theoretically discussed in \citet{kanagawa_AISTATS2014}. 
It is also worth noting that with similar probability estimate, \citet{pmlr-v89-hsu19a} constructed a marginal likelihood-like objective for hyperparameter selection of CMEs, within the context of likelihood-free inference. While they prove that this objective provides an asymptotically correct likelihood surrogate, it is not guaranteed to be positive or normalized for finite data. Consequently, we treat it solely as a proxy for conditional probability, limiting its use to hyperparameter selection criteria. 

We emphasize that our focus is on CMEs, i.e. on representing conditional distributions as mean elements within the RKHS. This enables two key advantages: first, we can represent distributions even without constraining the NN output, and second, it opens up applications beyond standard density estimation tasks, as showcased by the unique properties of KMEs explored in \cref{sec:RL}.

\subsubsection{Approach 1: Iterative Optimization}\label{Approach1}
Based on the conditional probability estimate \eqref{eq:cond_density}, we propose optimizing the bandwidth parameter $\sigma$ to minimize an objective function based on $L^2$ norm. Accordingly, we adopt the following squared (SQ) error loss:
\begin{align*}
\mathcal{L}_{SQ}=\frac{1}{2}\iint_{}^{}\left(\hat{p}(y|x)-p(y|x)\right) ^2p(x)dxdy.
\end{align*}
This objective function has been used in the density-ratio-based conditional density estimation method \citep{pmlr-v9-sugiyama10a}. In our case, by plugging in $\hat{p}(y|x)=\sum_{a=1}^{M} k_{\sigma}(y, \eta_{a}) f_a(x;\theta)$, we obtain the following empirical loss $\frac1n \sum_{i=1}^n \hat{\ell}_{SQ}(\sigma)$, where:
\begin{align*}
\hat{\ell}_{SQ}(\sigma)= -2\sum_{a}k_{\sigma}(y_i,\eta_a)w_a+\sum_{a,b}k_{\sqrt{2}\sigma}(\eta_a,\eta_b)w_aw_b.
\end{align*}
The derivation is given in \cref{subsec:SQ_derivation}. In practice, we iteratively optimize $\theta$ and $\sigma$, through $\min_{\theta}\frac1n \sum_{i=1}^n \hat{\ell}(\theta)$ and $\min_{\sigma}\frac1n \sum_{i=1}^n \hat{\ell}_{SQ}(\sigma)$, every step. Since $\eta_a$ within $k_{\sigma}(\cdot,\eta_a)$ are model parameters, and not randomly sampled during optimization, $\sigma$ can be optimized stably using minibatch optimization, with minimal gradient variance. Complemented with regularization techniques such as weight decay \cite{loshchilov2018decoupled}, we observe that the optimization can generally be carried out without encountering severe overfitting. 

\subsubsection{Approach 2: Joint Optimization}\label{Approach2}
Interestingly, the only distinction between \eqref{eq:RKHS_loss} and $\hat{\ell}_{SQ}(\sigma)$ lies in their second terms, representing the squared functional norms on $\calH_{\sigma}$ and $\calH_{\sqrt{2}\sigma}$, respectively. This close resemblance hints at the potential for joint optimization of $\theta$ and $\sigma$ using the objective function given in \eqref{eq:RKHS_loss}, expressed as $\min_{\theta,\sigma}\frac1n \sum_{i=1}^n \hat{\ell}(\theta, \sigma)$. The following theorem provides a key justification for this approach by establishing an inequality that connects these two norms:
\begin{theorem}{}
\label{theorem:rkhs_norms}
Let $f= \sum_{a=1}^{M} k_{\sqrt{2}\sigma}(\cdot, \eta_{a})w_a\in\calH_{\sqrt{2}\sigma}$ and $g=\sum_{a=1}^{M} k_{\sigma}(\cdot, \eta_{a})w_a\in\calH_{\sigma}$. Then the following inequality holds:
\begin{align*}
\|f\|_{\calH_{\sqrt{2}\sigma}}\leq \|g\|_{\calH_{\sigma}}.
\end{align*}
\end{theorem}
The proof, provided in \cref{subsec:Proof_Kenji}, leverages the Fourier transform expression of RKHS for translation invariant kernels. By replacing the second term of $\hat{\ell}_{SQ}(\sigma)$ with this upper bound, we observe a coincidence with the original objective function \eqref{eq:RKHS_loss}. This result validates our proposal for \emph{joint} optimization of $\theta$ and $\sigma$ using \eqref{eq:RKHS_loss}, since it simultaneously optimizes an upper bound of $\hat{\ell}_{SQ}(\sigma)$ for $\sigma$, while optimizing the original loss function for $\theta$. Joint optimization not only simplifies the procedure but also, as a valuable byproduct, mitigates overfitting in hyperparameter optimization due to the utilization of an upper bound. Indeed, we observe that joint optimization, with its combined advantages, yields stable performance across various datasets.

\subsection{Related Work}
Mixture Density Networks (MDNs) \citep{bishop1994mixture} model conditional probability distributions as a mixture of (typically) Gaussian components, where the NN directly predicts means, variances, and mixing weights. Due to their simplicity, MDNs have been used in diverse tasks \cite{papamakarios2016fast, MICB19}. 

Normalizing Flows (NFs) \citep{rezende2015variational, papamakarios2021normalizing} transform a simple base distribution using invertible mappings parameterized by NNs, ultimately yielding the desired distribution. Particularly, conditional version of NFs that utilize monotonic rational-quadratic splines for the transformation \citep{durkan2019neural}, have found extensive use in simulation-based inference settings \citep{lueckmann2021benchmarking}. 

Also, \citet{han2022card} recently proposed CARD, a conditional density estimator (and classifier) based on diffusion models. CARD combines a denoising diffusion-based conditional generative model with a pre-trained conditional mean estimator, achieving strong performance in diverse practical settings.

We will compare these methods with our proposed approaches in the following experimental sections. 

\section{Experiments on Density Estimation}\label{sec:experiment}
To investigate the effectiveness of our approach, we conduct experiments on both toy and real-world datasets. We focus on conditional density estimation tasks where the dimension of the output variable is one ($d_y=1$). Note that the input variable can be multi-dimensional, and we conduct experiments on such settings using UCI datasets \cite{uci_mldataset}, where $d_x$ can be up to 90.

We denote our approach proposed in Subsections \ref{Approach1} and \ref{Approach2} as Proposal-Iterative and Proposal-Joint, respectively. We compare with DF, MDN, NF, and CARD. For DF, we tested two models: DF used with the median heuristic for bandwidth selection (DF-med) and DF with bandwidth fixed to 0.1 (DF-0.1). For NF, we use the conditional version of autoregressive Neural Spline Flow \citep{durkan2019neural}. 

Both evaluation metrics employed in these settings require samples from the learned model. To sample points from CME-based approaches, we employ kernel herding \citep{chen2010super}, a deterministic sampling method that yields super-samples.

\begin{table*}
\caption{\label{tab:WAS}WAS1 for toy datasets. Values are multiplied by $100$.}\vspace{1mm}
\begin{center}
\resizebox{15cm}{!}{
\begin{tabular}{@{}l|ccccccc@{}}
\toprule[1.5pt]
Dataset  &                     &                     & WAS1 $\downarrow$                &                     &                     \\
         & Proposal-Iterative                 & Proposal-Joint          & DF-med  &DF-0.1     & MDN     &NF               & CARD              \\ \midrule
Bimodal   & $5.88\pm 0.28$  & $\bm{5.67\pm 0.28}$  & $10.87\pm 0.23$  & $5.93\pm 0.23$ & $6.23\pm 0.27$  & $6.83\pm 0.50$ & $6.59\pm 0.46$  \\
Skewed & $4.86\pm 0.23$  & $\bm{4.68\pm 0.22}$  & $6.07\pm 0.19$  & $5.26\pm 0.19$ & $5.76\pm 0.18$  & $6.56\pm 0.29$ & $5.93\pm 0.26$  \\
Ring   & $\bm{21.13\pm 0.99}$ & $\bm{21.10\pm 1.14}$  & $26.17\pm 1.16$  & $22.16\pm 1.25$ & $26.66\pm 0.80$   & $27.99\pm 1.74$ & $29.91\pm 0.97$\\ \bottomrule[1.5pt]
\end{tabular}
}
\end{center}
\end{table*}

\begin{table*}
\caption{\label{tab:QICE}QICE (in ${\%}$) for UCI datasets.}\vspace{1mm}
\begin{center}
\resizebox{15cm}{!}{
\begin{tabular}{@{}l|ccccccc@{}}
\toprule[1.5pt]
Dataset  &                     &                     & QICE $\downarrow$                &                     &                     \\
         & Proposal-Iterative                 & Proposal-Joint        & DF-med  & DF-0.1      & MDN     &NF               & CARD              \\ \midrule
Boston   & $\bm{3.15\pm 0.92}$  & $\bm{3.09\pm 0.53}$  & $5.74\pm 1.17$ & $3.88\pm 0.82$  & $4.85\pm 1.19$  & $4.09\pm 0.95$ & $3.45\pm 0.83$  \\
Concrete & $3.06\pm 0.82$  & $3.21\pm 0.72$ & $4.05\pm 0.95$  & $4.11\pm 0.66$  & $3.21\pm 0.79$  & $2.90\pm 0.79$ & $\bm{2.30\pm 0.66}$  \\
Energy   & $\bm{2.89\pm 0.69}$ & $3.29\pm 0.73$ & $7.14\pm 0.88$  & $3.23\pm 0.90$  & $3.54\pm 0.85$   & $3.62\pm 1.23$ & $4.91\pm 0.94$  \\
Kin8nm   & $0.98\pm 0.29$ & $\bm{0.91\pm 0.19}$ & $4.70\pm 0.32$ & $2.60\pm 0.41$ & $2.42\pm 0.32$  & $1.75\pm 0.87$ & $\bm{0.92\pm 0.25}$ \\
Naval    & $10.88\pm 1.09$  & $6.81\pm 1.07$ & $7.09\pm 1.04$ & $8.71\pm 0.37$ & $8.34\pm 3.41$  & $4.41\pm 1.54$ & $\bm{0.80\pm 0.21}$ \\
Power    & $0.88\pm 0.24$  & $\bm{0.84\pm 0.18}$ & $4.81\pm 0.33$ & $2.91\pm 0.18$  & $1.34\pm 0.35$   & $1.35\pm 0.59$ & $0.92\pm 0.21$  \\
Protein  & $\bm{0.48\pm 0.05}$  & $0.55\pm 0.17$ & $1.83\pm 0.19$  & $0.80\pm 0.07$  & $1.09\pm 0.35$  & $0.80\pm 0.38$  & $0.71\pm 0.11$  \\
Year     & $0.71\pm$ NA       & $\bm{0.53\pm}$ NA      & $1.80\pm$ NA      & $0.56\pm$ NA       & $0.74\pm$ NA       & $1.02\pm$ NA  & $\bm{0.53\pm}$ NA         \\ \bottomrule[1.5pt]
\end{tabular}
}
\end{center}
\end{table*}

\subsection{Toy Data}\label{exp:toy}
{\bf Set Up:} 

We conduct experiments on three distinct toy datasets: Bimodal, Skewed, and Ring. Each dataset features a one-dimensional input variable and comprises 5000 data points for training. 

These datasets are designed to challenge models with diverse distributional characteristics: Bimodal dataset comprised of two Gaussian distributions, but the degree of bimodality and noise levels vary depending on the value of $x$. Skewed dataset are sampled from a skew-normal distribution, but distribution parameters, such as skewness, change as a function of $x$. Ring dataset features a ring-shaped distribution with an embedded box-shaped distribution. The detailed generation process as well as a figure for each dataset is provided in \cref{toy_details}.

{\bf Evaluation Metric:} 

We evaluate models using a metric based on the Wasserstein-1 distance (WAS1), defined by 
\begin{align*}
W_1(\mu, \nu) = \inf_{\pi \in \Gamma(\mu, \nu)} \int_{\mathbb{R}\times\mathbb{R}} |x - y| d\pi(x, y), 
\end{align*}
where $\Gamma(\mu, \nu)$ is the set of probability distributions whose marginals are $\mu$ and $\nu$. In the one-dimensional case, this metric can be readily calculated using packages like SciPy \citep{2020SciPy-NMeth}.

The evaluation process involves the following steps: we first draw 50 samples from the learned models for each 200 equally spaced evaluation points $x_{\text{test}}$. We then, compute the WAS1s between these samples and 50 points sampled from the true generating process (for each $x_{\text{test}}$). We average the WAS1 values for all evaluation points and report the mean and standard deviation across 10 independent runs.

{\bf Results:} 

The overall results are presented in \cref{tab:WAS}, demonstrating that our proposed approaches outperform competitors including NF and CARD. Crucially for DF, the median heuristic fails to achieve optimal performance. It only becomes competitive with our approaches when the parameter is set to 0.1, which was identified through empirical trials. This highlights the general challenge of hyperparameter selection for the output variable kernel in CME-based methods, where both our iterative and joint approaches demonstrably lead to more effective parameter selection. 

\subsection{UCI Datasets}\label{exp:uci}
{\bf Set Up:}

To further investigate our approaches, we conduct experiments on 8 real-world regression benchmark datasets from the UCI repository. Details of the datasets are provided in \cref{uci_details}.

We follow experimental protocols of \citet{han2022card}: (a) we employ the same train-test splits with a $90\%/10\%$ ratio, and use 20 folds for all datasets except Protein (5 folds) and Year (1 fold), (b) we standardize both input and output variables for training and remove standardization for evaluation,  and (c) for each test data point $x_{\text{test}}$, we sample 1000 points from learned models, conditioned on $x_{\text{test}}$, and calculate the metric described below.

{\bf Evaluation Metric:} 

Since UCI datasets do not have ``true generating process", we adopt the Quantile Interval Coverage Error (QICE) metric proposed also by \citet{han2022card}. To compute QICE, we first generate sufficient number of samples for each conditional values $x_i$. We then divide the generated samples into $L$ equally spaced bins, resulting in $L$ quantile intervals with boundaries denoted as $\hat{y}_i^{\text{low}_j}$ and $\hat{y}_i^{\text{high}_j}$. Then compute the following quantity:
\begin{align*}
    \text{QICE}\coloneqq \frac{1}{L}\sum_{j=1}^L\bigg|r_j-\frac{1}{L}\bigg|, 
\text{ where }\\
    r_j = \frac{1}{n}\sum_{i=1}^n{1}_{y_i\geq\hat{y}_i^{\text{low}_j}}\cdot{1}_{y_i\leq\hat{y}_i^{\text{high}_j}}. 
\end{align*}

When the learned conditional distribution perfectly matches the true distribution, we expect approximately $1/L$ of the true data to fall within each of the $L$ quantile intervals, resulting in a QICE value of 0. In this experiment, we follow \citet{han2022card} and set $L=10$. We report mean and standard deviation of the QICE metric across all splits.

{\bf Results:} 

The overall results are presented in \cref{tab:QICE}. It can be seen that our proposed approaches frequently outperform existing methods such as DF, MDN, and NF, and often prove competitive with CARD in terms of the QICE metric. This is particularly impressive considering CARD requires two separate NN optimization procedures: 1. Conditional mean estimator optimization, and 2. Denoising diffusion model optimization, guided by the optimized conditional mean estimator. In contrast, our approaches achieve a comparable level of high-quality density estimation as diffusion-based CARD, while only using a single NN. The Proposal-Joint approach stands out with its exceptional efficiency, requiring only a standard NN training procedure.

It is also worth noting that in the UCI experiments, DF-0.1 clearly underperforms compared to our approaches, crucially demonstrating the importance of tailoring bandwidth $\sigma$ to each dataset for optimal performance. This suggests that a fixed $\sigma$ value, as used in DF-0.1, is typically insufficient for achieving good results across diverse datasets.

\section{Applications to Reinforcement Learning}\label{sec:RL}
\begin{figure*}[ht]
\vskip 0.2in
\begin{center}
\centerline{\includegraphics[width=1.0\textwidth]{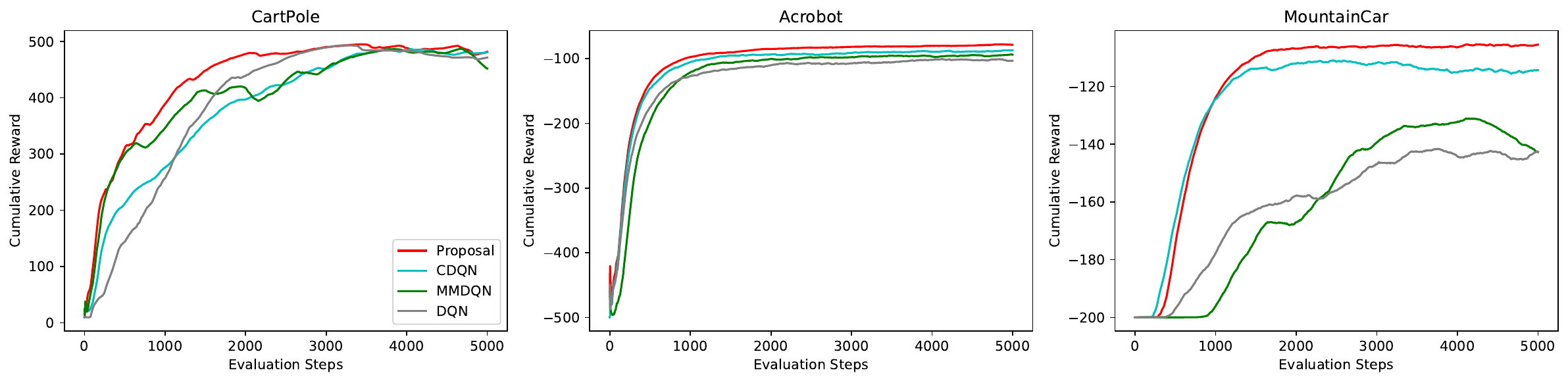}}
\caption{Performance comparison on three environments. We report the mean of cumulative rewards across 10 independent runs.}
\label{RL_rewards}
\end{center}
\vskip -0.2in
\end{figure*}
\subsection{Backgrounds}
We consider a classical setting in RL, where the framework of a Markov Decision Process (MDP) \citep{puterman2014markov} governs the agent-environment interactions. MDP is defined by the tuple $(\mathcal{S},\mathcal{A},R,P,\gamma)$, where $\mathcal{S}$ denotes the state space, $\mathcal{A}$ the action space, $P(\cdot|s,a)$ the transition probabilities, $R(s,a)$ the reward dependent on the state and action, and $\gamma\in\left(0,1\right]$ the discount factor. We represent the discounted sum of rewards received by an agent under a policy $\pi$ as a random variable $Z^{\pi}(s,a)=\sum_{t=0}^{\infty}\gamma^{t}R(s_t,a_t)$ on space $\mathcal{Z}$, where $s_0=s$, $a_0=a$, $s_{t+1}\sim P(\cdot|s_t,a_t)$, and $a_t\sim \pi(\cdot|s_t)$. 

The state-action value, also known as Q-value, is defined as $Q^{\pi}(s,a)=\expect{Z^{\pi}(s,a)}$. In Q-learning \citep{watkins1992qlearning}, the goal is to learn the optimal Q-value, $Q^{\star}(s,a)=\max_\pi Q^{\pi}(s,a)$, which is the fixed point of the Bellman optimality operator $\mathcal{T}$: 
\begin{align*}
(\mathcal{T}Q)(s,a) \equiv \expect{R(s,a)}+\gamma\expect[P]{\max_{a^{\prime}} Q^{}(s^{\prime},a^{\prime})},
\end{align*}
with $s'\sim P(\cdot|s,a)$.  Deep Q-Network (DQN) \citep{mnih2015human} represents Q-values with NNs, parameterized by $\theta$, and optimizes them based on the following loss function:
\begin{align*}
\hat{\ell}(\theta)=\left(r+\gamma \max_{a^{\prime}}Q_{\theta^{-}}(s^{\prime},a^{\prime})-Q_{\theta}(s,a)\right)^2,
\end{align*}
where $\theta^{-}$ corresponds to the fixed target network, periodically copied from and synchronized with $Q_\theta$. A large replay buffer stores experienced transitions $(s,a,r,s^{\prime})$, and batches are sampled for mini-batch optimization, consistent with standard NN training practice.

Instead of only learning the expectation of $Z(s,a)$, distributional RL (DRL) \citep{bellemare2017distributional,bdr2023} aims to approximate its entire distribution. Accordingly, distributional Bellman optimality operator can be defined:
\begin{align*}
(\mathcal{T}Z)(s,a) \overset{D}\equiv R(s,a)+\gamma Z\left(s^{\prime},\argmax_{a^{\prime}}\expect[P]{Z^{}(s^{\prime},a^{\prime})}\right)
\end{align*}
where, $X\overset{D}\equiv Y$ indicates that random variables $X$ and $Y$ follow the same distribution. 

Several DRL approaches have significantly advanced the field of RL, achieving performance gains beyond standard DQN. One such approach, Categorical DQN (CDQN) \citep{bellemare2017distributional}, models distributions as categorical distribution represented by $\sum_{a=1}^M\theta_a\delta_a$, where $\theta_a$ are learnable parameters and $\delta_a$ represent fixed discrete atoms on a pre-defined grid. Alternatively, some approaches utilize quantile regression for  distribution modeling \cite{dabney2018distributional}. More recently, \citet{Nguyen-Tang_Gupta_Venkatesh_2021} proposed Moment Matching DQN (MMDQN), which represents distributions in RKHS. While \citet{Nguyen-Tang_Gupta_Venkatesh_2021} did not explicitly formulate their approach based on CMEs, their formulation can be interpreted as a CME variant with uniform weights. We compare and contrast these approaches with our proposed method in \cref{sec:RL_Discussion}. 
\subsection{Proposed Model for DRL}
We propose modeling the (conditional) distribution of $Z(s,a)$ using our NN-CME hybrid approach. This approach leverages a kernel $k_\calZ(\cdot, z)\in\calH_\calZ$ and represents distributions as:
\begin{align*}
\mu_{P(Z|S,A)}(x) = \sum_{a=1}^{M} k_\calZ(\cdot, \eta_a)f_a(x;\theta),
\end{align*}
where $x$ is a tuple of state and action. Here, $\eta_a$ are chosen analogously to atoms $\delta_a$ in CDQN, effectively covering anticipated support of the distribution. Importantly, our CME parameterization excels by bypassing the need for matrix inversion, enabling efficient action selection. In contrast, applying existing CME approaches (such as DF) in this setting would necessitate evaluating $\vec{\beta}(x)$ in \eqref{eq:CME_empirical} by accessing the entire replay buffer and performing matrix inversion on this data. This can be prohibitively expensive, especially when performed repeatedly.

To construct a loss function in the context of deep Q-learning, we define a metric between the distribution of $R(s,a)+\gamma Z\left(s^{\prime},a^{\prime}\right)$ and that of $Z(s,a)$. Maximum Mean Discrepancy (MMD) \citep{Gretton2012MMD} provides a principled way to measure the discrepancy between these distributions in the RKHS:
\begin{align*}
\hat{d}_{k_{\calZ}} = \left\Vert \sum_{a=1}^{M} k_\calZ(\cdot, \tau_a)v_a -\sum_{a=1}^{M} k_\calZ(\cdot, \eta_a)w_a \right\Vert_{\calH_\calZ},
\end{align*}
where $\tau_a=r+\gamma \eta_a$ and $v_a=f_a(x;\theta^{-})$. Squaring this MMD leads to our DRL-specific loss function $\hat{d}_{k_{\calZ}}^2(\theta)$:
\begin{align*}
\hat{d}^2_{k_{\calZ}}(\theta) &= \sum_{a,b}k_\calZ(\tau_a,\tau_b)v_av_b-2\sum_{a,b}k_\calZ(\tau_a,\eta_b)v_aw_b\\
&+\sum_{a,b}k_\calZ(\eta_a,\eta_b)w_aw_b.
\end{align*}

Thus, we have retained the CME parametrization from \cref{sec:proposed-method}, but crucially adopted an MMD-based loss function to suit RL tasks. This flexible adaptation is enabled by both representing distributions as mean elements in the RKHS and exploiting the unique property of KMEs.

\subsection{Fusing Kernels}
Selecting the hyperparameters for $k_\calZ$ is an important issue, also in RL settings. Corollary 4.1 of \citet{killingberg2023the} implies that careful selection of the bandwidth parameter is crucial for ensuring theoretical convergence in moment matching-based DRL, when used with the Gaussian kernel. Inspired by MMD-FUSE \citep{biggs2023mmdfuse}, an MMD-based two-sample testing approach, we propose using a distribution over kernels, $k\in\mathcal{K}$. Our approach employs the following log-sum-exp type loss function:

\begin{align*}
\text{FUSE} = \log\left(\expect[k\sim\omega]{\exp({\hat{d}^2_{k_{\calZ}}(\theta)})}\right)
\end{align*}
where $\omega$ is an element of  $\mathcal{M}(\mathcal{K})$, the set  of distributions over the kernels. For instance, we can define $\omega$ as a uniform distribution over collections of Gaussian kernels with various bandwidths $\sigma>0$, where $\sigma$ are chosen from a uniformly discretized interval. The Donsker-Varadhan equality \citep{donsker1975asymptotic}, as stated in \citet{biggs2023mmdfuse}, offers the following interpretation: $\text{FUSE} = \sup_{\rho}\expect[k\sim\rho]{\hat{d}_k(\theta)}-\text{KL}[\rho\|\omega]$,
where KL denotes the  Kullback-Leibler divergence, $\rho\in\mathcal{M}(\mathcal{K})$ can be loosely interpreted as the ``posterior'' over kernels, and $\omega$ as a ``prior''. While a theoretical analysis of this approach within the DRL context remains important future work, empirical results suggest its performance advantages compared to using a single fixed kernel. 

\subsection{Discussions}\label{sec:RL_Discussion}
Both our approach and MMDQN utilize CMEs to model the distributions of $Z(s,a)$. However, they differ in what their NNs parameterize. MMDQN directly learns atoms through its network and assigns them uniform weights, resulting in the representation: $\mu_{P(Z|S,A)}(x) = \frac1M\sum_{a=1}^{M} k_{\calZ}(g_a(x;\theta),\cdot)$. While this eliminates the need for manual atom specification, its reliance on uniform weights might hinder its ability to accurately model complex distributions exhibiting multimodality or skewness.

When the predefined atoms $\eta_a$ in our approach coincide with the fixed atoms $\delta_a$ in CDQN, the parameterizations closely resemble each other. However, a key difference arises in the loss functions: CDQN employs cross-entropy loss, treating the distribution as categorical, whereas we leverage an MMD-based loss function. In essence, while our CME-based representation and MMD-based loss share similarities with MMDQN, our approach aligns more closely with CDQN in its atom structure and NN parameterization, but employs a distinct loss function.

\subsection{Experimental Results}
{\bf Set Up:} 

To investigate the unique aspects of our method and gain insights into DRL design choices, we tested it on three classic control environments from Gymnasium \cite{towers_gymnasium_2023}: CartPole, Acrobot, and MountainCar. While our approach does not inherently constrain the output of $f(x;\theta)$ to be positive or sum to one, we applied a softmax function in the last layer, analogous to CDQN, making the NN architecture identical to that of CDQN.

We note that while DRL algorithms are often evaluated on complex Atari2600 games, such experiments require extensive computational resources. For instance, the work by \citet{obando2020revisiting} required approximately 5 days to run each agent on each Atari environment using a GPU. As \citet{obando2020revisiting} demonstrated through comprehensive benchmark evaluations, valuable scientific insights can still be drawn from smaller-scale environments. Therefore, our focus in this experiment is on understanding how different parameterizations and loss functions impact various aspects of learning and control effectiveness.

{\bf Results:} 

\cref{RL_rewards} presents the overall results, demonstrating the consistent effectiveness of our approach across three environments. In CartPole, all methods achieve near-optimal performance, attaining approximately 500 cumulative reward. However, our method demonstrates faster and more stable convergence. In Acrobot, while all methods exhibit stability, ours achieves the highest cumulative reward. On MountainCar, our approach clearly outperforms the others.

Importantly, despite identical distribution parameterization to CDQN, our approach consistently achieves superior performance. This suggests that the MMD-based loss function combined with our fusion strategy contributes to the improved performance. Investigating theoretical justification for this improvement may present a potentially fruitful research direction. 

Despite improved performance reported by \citet{Nguyen-Tang_Gupta_Venkatesh_2021} on Atari tasks, MMDQN displayed instability in classic control environments. We hypothesize that this disparity arises from the inherent differences in reward structures. Sparse rewards in Atari environments potentially favor MMDQN's direct particle placement learning, while predetermined atom structure might be more effective in classic 
control environments with denser rewards. This suggests that the choice of distribution parameterization may depend on the reward structure of environments, highlighting the influence of inductive bias. 

\section{Conclusion}
This paper introduced a new NN-CME hybrid that effectively addresses the key challenges of existing CME methods in terms of scalability, expressiveness, and hyperparameter selection. We proposed to replace the Gram matrix inversion needed for CME estimation with an expressive NN, and presented strategies for optimizing the hyperparameter of $k_y$ by leveraging the Gaussian density kernel. We demonstrated its effectiveness in representing conditional distributions: in density estimation tasks, it achieves competitive results, and in RL applications, the proposed method outperforms competing approaches in terms of cumulative reward.

Our approach seamlessly lends itself to diverse other settings, including causal inference tasks where using CMEs offers distinct advantages \cite{Singh2019KIV,MuaKanSaeMar21,ChaTonGonTehSej2021,ParShaSchMua21}. Furthermore, while our approach effectively handled one-dimensional outputs, verifying the effectiveness with more challenging multidimensional settings presents an important area for future investigation. Drawing inspiration from inducing points approaches in Gaussian Process literature \cite{hensman2013gaussian}, optimizing the location parameters $\eta_a$ could prove key to achieving effective performance in these broader landscapes.

\section*{Impact Statements}
This paper makes a contribution to general purpose machine learning methodology, by improving the practicability of models for representing conditional distributions. Such models have a broad applicability and a potential for impact as they are important ingredients in a wide range of learning tasks. When applying the proposed methodology, it is crucial to consider its potential limitations. For instance, our approach, similar to existing density estimation methods, does not readily capture inherent causal relationships between input and output variables, which could be critical for informed decision-making. We encourage practitioners to consider these limitations and use our approach carefully for making reliable decisions in relevant scenarios.




\nocite{langley00}

\bibliography{reference}

\begin{thebibliography}{52}
\providecommand{\natexlab}[1]{#1}
\providecommand{\url}[1]{\texttt{#1}}
\expandafter\ifx\csname urlstyle\endcsname\relax
  \providecommand{\doi}[1]{doi: #1}\else
  \providecommand{\doi}{doi: \begingroup \urlstyle{rm}\Url}\fi

\bibitem[Bellemare et~al.(2017)Bellemare, Dabney, and Munos]{bellemare2017distributional}
Bellemare, M.~G., Dabney, W., and Munos, R.
\newblock A distributional perspective on reinforcement learning.
\newblock In \emph{International Conference on Machine Learning}, pp.\  449--458. PMLR, 2017.

\bibitem[Bellemare et~al.(2023)Bellemare, Dabney, and Rowland]{bdr2023}
Bellemare, M.~G., Dabney, W., and Rowland, M.
\newblock \emph{Distributional Reinforcement Learning}.
\newblock MIT Press, 2023.
\newblock \url{http://www.distributional-rl.org}.

\bibitem[Biggs et~al.(2023)Biggs, Schrab, and Gretton]{biggs2023mmdfuse}
Biggs, F., Schrab, A., and Gretton, A.
\newblock {MMD-FUSE}: {L}earning and combining kernels for two-sample testing without data splitting.
\newblock 2023.
\newblock URL \url{https://arxiv.org/abs/2306.08777}.

\bibitem[Bishop(1994)]{bishop1994mixture}
Bishop, C.~M.
\newblock Mixture density networks.
\newblock Report NCRG/94/004, 1994.

\bibitem[Chau et~al.(2021)Chau, Ton, Gonzalez, Teh, and Sejdinovic]{ChaTonGonTehSej2021}
Chau, S.~L., Ton, J.-F., Gonzalez, J., Teh, Y.~W., and Sejdinovic, D.
\newblock {BayesIMP: Uncertainty Quantification for Causal Data Fusion}.
\newblock In \emph{Advances in Neural Information Processing Systems (NeurIPS)}, volume~34, pp.\  3466--3477, 2021.

\bibitem[Chen et~al.(2010)Chen, Welling, and Smola]{chen2010super}
Chen, Y., Welling, M., and Smola, A.
\newblock Super-samples from kernel herding.
\newblock In \emph{Proceedings of the 26th Conference on Uncertainty in Artificial Intelligence}, pp.\  109--116. AUAI Press, 2010.

\bibitem[Dabney et~al.(2018)Dabney, Rowland, Bellemare, and Munos]{dabney2018distributional}
Dabney, W., Rowland, M., Bellemare, M., and Munos, R.
\newblock Distributional reinforcement learning with quantile regression.
\newblock In \emph{Proceedings of the AAAI Conference on Artificial Intelligence}, volume~32, 2018.

\bibitem[Donsker \& Varadhan(1975)Donsker and Varadhan]{donsker1975asymptotic}
Donsker, M.~D. and Varadhan, S. R.~S.
\newblock Asymptotic evaluation of certain markov process expectations for large time.
\newblock \emph{Communications on Pure and Applied Mathematics}, 28\penalty0 (1):\penalty0 1--47, 1975.

\bibitem[Dua \& Graff(2017)Dua and Graff]{uci_mldataset}
Dua, D. and Graff, C.
\newblock Uci machine learning repository, 2017.
\newblock URL \url{http://archive.ics.uci.edu/ml}.

\bibitem[Durkan et~al.(2019)Durkan, Bekasov, Murray, and Papamakarios]{durkan2019neural}
Durkan, C., Bekasov, A., Murray, I., and Papamakarios, G.
\newblock Neural spline flows.
\newblock In \emph{Advances in Neural Information Processing Systems}, pp.\  7509--7520, 2019.

\bibitem[Fukumizu et~al.(2004)Fukumizu, Bach, and Jordan]{fukumizu2004dimensionality}
Fukumizu, K., Bach, F.~R., and Jordan, M.~I.
\newblock Dimensionality reduction for supervised learning with reproducing kernel hilbert spaces.
\newblock In \emph{Journal of Machine Learning Research}, volume~5, pp.\  73--99, 2004.

\bibitem[Fukumizu et~al.(2008)Fukumizu, Gretton, Sun, and Schölkopf]{fukumizu2008kernel}
Fukumizu, K., Gretton, A., Sun, X., and Schölkopf, B.
\newblock Kernel measures of conditional dependence.
\newblock In \emph{Advances in Neural Information Processing Systems}, pp.\  489–496, 2008.

\bibitem[Fukumizu et~al.(2013)Fukumizu, Song, and Gretton]{Fukumizu2013KBR}
Fukumizu, K., Song, L., and Gretton, A.
\newblock Kernel {B}ayes' rule: {B}ayesian inference with positive definite kernels.
\newblock \emph{Journal of Machine Learning Research}, 14\penalty0 (82):\penalty0 3753--3783, 2013.

\bibitem[Girosi et~al.(1995)Girosi, Jones, and Poggio]{Girosi1995}
Girosi, F., Jones, M., and Poggio, T.
\newblock Regularization theory and neural networks architectures.
\newblock \emph{Neural Computation}, 7\penalty0 (2):\penalty0 219--269, 1995.
\newblock \doi{10.1162/neco.1995.7.2.219}.

\bibitem[Gretton et~al.(2005)Gretton, Herbrich, Smola, Schölkopf, and Hyvärinen]{gretton2005kernel}
Gretton, A., Herbrich, R., Smola, A., Schölkopf, B., and Hyvärinen, A.
\newblock Kernel methods for measuring independence.
\newblock In \emph{Journal of Machine Learning Research}, volume~6, pp.\  2075–2129, 2005.

\bibitem[Gretton et~al.(2012)Gretton, Borgwardt, Rasch, Sch{{\"o}}lkopf, and Smola]{Gretton2012MMD}
Gretton, A., Borgwardt, K.~M., Rasch, M.~J., Sch{{\"o}}lkopf, B., and Smola, A.
\newblock A kernel two-sample test.
\newblock \emph{Journal of Machine Learning Research}, 13\penalty0 (25):\penalty0 723--773, 2012.

\bibitem[Grünewälder et~al.(2012)Grünewälder, Lever, Baldassarre, Patterson, Gretton, and Pontil]{grunewaelder2012conditional}
Grünewälder, S., Lever, G., Baldassarre, L., Patterson, S., Gretton, A., and Pontil, M.
\newblock Conditional mean embeddings as regressors.
\newblock In \emph{Proceedings of the 29th International Conference on Machine Learning}, pp.\  1823--1830. Omnipress, 2012.

\bibitem[Han et~al.(2022)Han, Zheng, and Zhou]{han2022card}
Han, X., Zheng, H., and Zhou, M.
\newblock {CARD}: Classification and regression diffusion models.
\newblock In \emph{Advances in Neural Information Processing Systems}, 2022.

\bibitem[Hensman et~al.(2013)Hensman, Fusi, and Lawrence]{hensman2013gaussian}
Hensman, J., Fusi, N., and Lawrence, N.~D.
\newblock Gaussian processes for big data.
\newblock In \emph{Proceedings of the Conference on Uncertainty in Artificial Intelligence}, pp.\  282–290, 2013.

\bibitem[Hsu \& Ramos(2019)Hsu and Ramos]{pmlr-v89-hsu19a}
Hsu, K. and Ramos, F.
\newblock Bayesian learning of conditional kernel mean embeddings for automatic likelihood-free inference.
\newblock In \emph{Proceedings of the Twenty-Second International Conference on Artificial Intelligence and Statistics}, volume~89 of \emph{Proceedings of Machine Learning Research}, pp.\  2631--2640. PMLR, 2019.

\bibitem[Kanagawa \& Fukumizu(2014)Kanagawa and Fukumizu]{kanagawa_AISTATS2014}
Kanagawa, M. and Fukumizu, K.
\newblock {Recovering Distributions from Gaussian RKHS Embeddings}.
\newblock In Kaski, S. and Corander, J. (eds.), \emph{Proceedings of the Seventeenth International Conference on Artificial Intelligence and Statistics}, volume~33 of \emph{Proceedings of Machine Learning Research}, pp.\  457--465, Reykjavik, Iceland, 2014. PMLR.

\bibitem[Killingberg \& Langseth(2023)Killingberg and Langseth]{killingberg2023the}
Killingberg, L. and Langseth, H.
\newblock The multiquadric kernel for moment-matching distributional reinforcement learning.
\newblock \emph{Transactions on Machine Learning Research}, 2023.
\newblock ISSN 2835-8856.

\bibitem[Kim \& Scott(2012)Kim and Scott]{JMLR:v13:kim12b}
Kim, J. and Scott, C.~D.
\newblock Robust kernel density estimation.
\newblock \emph{Journal of Machine Learning Research}, 13\penalty0 (82):\penalty0 2529--2565, 2012.

\bibitem[Kingma \& Ba(2015)Kingma and Ba]{kingma2015adam}
Kingma, D.~P. and Ba, J.
\newblock Adam: A method for stochastic optimization.
\newblock In \emph{International Conference on Learning Representations}, 2015.

\bibitem[Loshchilov \& Hutter(2019)Loshchilov and Hutter]{loshchilov2018decoupled}
Loshchilov, I. and Hutter, F.
\newblock Decoupled weight decay regularization.
\newblock In \emph{International Conference on Learning Representations}, 2019.

\bibitem[Lueckmann et~al.(2021)Lueckmann, Boelts, Greenberg, Goncalves, and Macke]{lueckmann2021benchmarking}
Lueckmann, J.-M., Boelts, J., Greenberg, D., Goncalves, P., and Macke, J.
\newblock Benchmarking simulation-based inference.
\newblock In \emph{Proceedings of The 24th International Conference on Artificial Intelligence and Statistics}, volume 130 of \emph{Proceedings of Machine Learning Research}, pp.\  343--351. PMLR, 2021.

\bibitem[Makansi et~al.(2019)Makansi, Ilg, {\c{C}}i{\c{c}}ek, and Brox]{MICB19}
Makansi, O., Ilg, E., {\c{C}}i{\c{c}}ek, {\"O}., and Brox, T.
\newblock Overcoming limitations of mixture density networks: A sampling and fitting framework for multimodal future prediction.
\newblock In \emph{IEEE International Conference on Computer Vision and Pattern Recognition}, 2019.

\bibitem[Miyato et~al.(2018)Miyato, Kataoka, Koyama, and Yoshida]{miyato2018spectral}
Miyato, T., Kataoka, T., Koyama, M., and Yoshida, Y.
\newblock Spectral normalization for generative adversarial networks.
\newblock In \emph{International Conference on Learning Representations}, 2018.

\bibitem[Mnih et~al.(2015)Mnih, Kavukcuoglu, Silver, Rusu, Veness, Bellemare, Graves, Riedmiller, Fidjeland, Ostrovski, et~al.]{mnih2015human}
Mnih, V., Kavukcuoglu, K., Silver, D., Rusu, A.~A., Veness, J., Bellemare, M.~G., Graves, A., Riedmiller, M., Fidjeland, A.~K., Ostrovski, G., et~al.
\newblock Human-level control through deep reinforcement learning.
\newblock \emph{Nature}, 518\penalty0 (7540):\penalty0 529--533, 2015.

\bibitem[Muandet et~al.(2017)Muandet, Fukumizu, Sriperumbudur, Sch{\"o}lkopf, and Gretton]{muandet2017kernel}
Muandet, K., Fukumizu, K., Sriperumbudur, B., Sch{\"o}lkopf, B., and Gretton, A.
\newblock Kernel mean embedding of distributions: A review and beyond.
\newblock \emph{Foundations and Trends® in Machine Learning}, 10\penalty0 (1-2):\penalty0 1--141, 2017.

\bibitem[Muandet et~al.(2021)Muandet, Kanagawa, Saengkyongam, and Marukatat]{MuaKanSaeMar21}
Muandet, K., Kanagawa, M., Saengkyongam, S., and Marukatat, S.
\newblock Counterfactual mean embeddings.
\newblock \emph{Journal of Machine Learning Research}, 22\penalty0 (162):\penalty0 1--71, 2021.

\bibitem[Nguyen-Tang et~al.(2021)Nguyen-Tang, Gupta, and Venkatesh]{Nguyen-Tang_Gupta_Venkatesh_2021}
Nguyen-Tang, T., Gupta, S., and Venkatesh, S.
\newblock Distributional reinforcement learning via moment matching.
\newblock \emph{Proceedings of the AAAI Conference on Artificial Intelligence}, 35\penalty0 (10):\penalty0 9144--9152, 2021.

\bibitem[Obando-Ceron \& Castro(2021)Obando-Ceron and Castro]{obando2020revisiting}
Obando-Ceron, J.~S. and Castro, P.~S.
\newblock Revisiting rainbow: Promoting more insightful and inclusive deep reinforcement learning research.
\newblock In \emph{Proceedings of the 38th International Conference on Machine Learning}, Proceedings of Machine Learning Research. PMLR, 2021.

\bibitem[Papamakarios \& Murray(2016)Papamakarios and Murray]{papamakarios2016fast}
Papamakarios, G. and Murray, I.
\newblock Fast $\epsilon$-free inference of simulation models with bayesian conditional density estimation.
\newblock In \emph{Advances in Neural Information Processing Systems}, volume~29, pp.\  1028--1036, 2016.

\bibitem[Papamakarios et~al.(2021)Papamakarios, Nalisnick, Rezende, Mohamed, and Lakshminarayanan]{papamakarios2021normalizing}
Papamakarios, G., Nalisnick, E., Rezende, D.~J., Mohamed, S., and Lakshminarayanan, B.
\newblock Normalizing flows for modeling and inference.
\newblock \emph{Journal of Machine Learning Research}, 22\penalty0 (57):\penalty0 1--64, 2021.

\bibitem[Park et~al.(2021)Park, Shalit, Sch{\"o}lkopf, and Muandet]{ParShaSchMua21}
Park, J., Shalit, U., Sch{\"o}lkopf, B., and Muandet, K.
\newblock Conditional distributional treatment effect with kernel conditional mean embeddings and u-statistic regression.
\newblock In \emph{Proceedings of 38th International Conference on Machine Learning}, volume 139 of \emph{Proceedings of Machine Learning Research}, pp.\  8401--8412. PMLR, 2021.

\bibitem[Puterman(2014)]{puterman2014markov}
Puterman, M.~L.
\newblock \emph{Markov decision processes: Discrete stochastic dynamic programming}.
\newblock John Wiley \& Sons, 2014.

\bibitem[Rezende \& Mohamed(2015)Rezende and Mohamed]{rezende2015variational}
Rezende, D.~J. and Mohamed, S.
\newblock Variational inference with normalizing flows.
\newblock In \emph{Proceedings of The 32nd International Conference on Machine Learning}, volume~37, pp.\  1530–1538. PMLR, 2015.

\bibitem[Saitoh(1997)]{Saitoh1997}
Saitoh, S.
\newblock \emph{Integral transforms, reproducing kernels and their applications}.
\newblock Pitman research notes in mathematics series ; 369. Chapman and Hall/CRC, 1997.

\bibitem[Schaul et~al.(2016)Schaul, Quan, Antonoglou, and Silver]{schaul2015prioritized}
Schaul, T., Quan, J., Antonoglou, I., and Silver, D.
\newblock Prioritized experience replay.
\newblock In \emph{International Conference on Learning Representations}, 2016.

\bibitem[Singh et~al.(2019)Singh, Sahani, and Gretton]{Singh2019KIV}
Singh, R., Sahani, M., and Gretton, A.
\newblock Kernel instrumental variable regression.
\newblock In \emph{Advances in Neural Information Processing Systems}, volume~32, pp.\  4595–4607, 2019.

\bibitem[Song et~al.(2009)Song, Huang, Smola, and Fukumizu]{Song2009CME}
Song, L., Huang, J., Smola, A., and Fukumizu, K.
\newblock Hilbert space embeddings of conditional distributions with applications to dynamical systems.
\newblock In \emph{International Conference on Machine Learning}, pp.\  961--968, 2009.

\bibitem[Song et~al.(2013)Song, Fukumizu, and Gretton]{song2013kernelembeddings}
Song, L., Fukumizu, K., and Gretton, A.
\newblock Kernel embeddings of conditional distributions: A unified kernel framework for nonparametric inference in graphical models.
\newblock \emph{IEEE Signal Processing Magazine}, 30\penalty0 (4):\penalty0 98--111, 2013.

\bibitem[Sriperumbudur et~al.(2010)Sriperumbudur, Gretton, Fukumizu, Sch\"{o}lkopf, and Lanckriet]{Sriperumbudur2010KME}
Sriperumbudur, B.~K., Gretton, A., Fukumizu, K., Sch\"{o}lkopf, B., and Lanckriet, G.~R.
\newblock Hilbert space embeddings and metrics on probability measures.
\newblock \emph{Journal of Machine Learning Research}, 11:\penalty0 1517–1561, 2010.

\bibitem[Stimper et~al.(2023)Stimper, Liu, Campbell, Berenz, Ryll, Schölkopf, and Hernández-Lobato]{Stimper2023}
Stimper, V., Liu, D., Campbell, A., Berenz, V., Ryll, L., Schölkopf, B., and Hernández-Lobato, J.~M.
\newblock normflows: A pytorch package for normalizing flows.
\newblock \emph{Journal of Open Source Software}, 8\penalty0 (86):\penalty0 5361, 2023.

\bibitem[Sugiyama et~al.(2010)Sugiyama, Takeuchi, Suzuki, Kanamori, Hachiya, and Okanohara]{pmlr-v9-sugiyama10a}
Sugiyama, M., Takeuchi, I., Suzuki, T., Kanamori, T., Hachiya, H., and Okanohara, D.
\newblock Conditional density estimation via least-squares density ratio estimation.
\newblock In \emph{Proceedings of the Thirteenth International Conference on Artificial Intelligence and Statistics}, volume~9 of \emph{Proceedings of Machine Learning Research}, pp.\  781--788. PMLR, 2010.

\bibitem[Towers et~al.(2023)Towers, Terry, Kwiatkowski, Balis, Cola, Deleu, Goulão, Kallinteris, KG, Krimmel, Perez-Vicente, Pierré, Schulhoff, Tai, Shen, and Younis]{towers_gymnasium_2023}
Towers, M., Terry, J.~K., Kwiatkowski, A., Balis, J.~U., Cola, G.~d., Deleu, T., Goulão, M., Kallinteris, A., KG, A., Krimmel, M., Perez-Vicente, R., Pierré, A., Schulhoff, S., Tai, J.~J., Shen, A. T.~J., and Younis, O.~G.
\newblock Gymnasium, 2023.
\newblock URL \url{https://zenodo.org/record/8127025}.

\bibitem[van Hasselt et~al.(2016)van Hasselt, Guez, and Silver]{vanhasselt2015doubledqn}
van Hasselt, H., Guez, A., and Silver, D.
\newblock Deep reinforcement learning with double q-learning.
\newblock In \emph{Proceedings of the Thirtieth AAAI Conference on Artificial Intelligence}, pp.\  2094--2100. AAAI Press, 2016.

\bibitem[Virtanen et~al.(2020)Virtanen, Gommers, Oliphant, Haberland, Reddy, Cournapeau, Burovski, Peterson, Weckesser, Bright, {van der Walt}, Brett, Wilson, Millman, Mayorov, Nelson, Jones, Kern, Larson, Carey, Polat, Feng, Moore, {VanderPlas}, Laxalde, Perktold, Cimrman, Henriksen, Quintero, Harris, Archibald, Ribeiro, Pedregosa, {van Mulbregt}, and {SciPy 1.0 Contributors}]{2020SciPy-NMeth}
Virtanen, P., Gommers, R., Oliphant, T.~E., Haberland, M., Reddy, T., Cournapeau, D., Burovski, E., Peterson, P., Weckesser, W., Bright, J., {van der Walt}, S.~J., Brett, M., Wilson, J., Millman, K.~J., Mayorov, N., Nelson, A. R.~J., Jones, E., Kern, R., Larson, E., Carey, C.~J., Polat, {\.I}., Feng, Y., Moore, E.~W., {VanderPlas}, J., Laxalde, D., Perktold, J., Cimrman, R., Henriksen, I., Quintero, E.~A., Harris, C.~R., Archibald, A.~M., Ribeiro, A.~H., Pedregosa, F., {van Mulbregt}, P., and {SciPy 1.0 Contributors}.
\newblock {{SciPy} 1.0: Fundamental Algorithms for Scientific Computing in Python}.
\newblock \emph{Nature Methods}, 17:\penalty0 261--272, 2020.

\bibitem[Watkins \& Dayan(1992)Watkins and Dayan]{watkins1992qlearning}
Watkins, C.~J. and Dayan, P.
\newblock Q-learning.
\newblock \emph{Machine learning}, 8\penalty0 (3-4):\penalty0 279--292, 1992.

\bibitem[Xu et~al.(2021)Xu, Chen, Srinivasan, de~Freitas, Doucet, and Gretton]{xu2021learning}
Xu, L., Chen, Y., Srinivasan, S., de~Freitas, N., Doucet, A., and Gretton, A.
\newblock Learning deep features in instrumental variable regression.
\newblock In \emph{International Conference on Learning Representations}, 2021.

\bibitem[Xu et~al.(2022)Xu, Chen, Doucet, and Gretton]{pmlr-v162-xu22a}
Xu, L., Chen, Y., Doucet, A., and Gretton, A.
\newblock Importance weighted kernel {B}ayes’ rule.
\newblock In \emph{Proceedings of the 39th International Conference on Machine Learning}, volume 162 of \emph{Proceedings of Machine Learning Research}, pp.\  24524--24538. PMLR, 2022.

\end{thebibliography}
\bibliographystyle{icml2024}

\newpage
\appendix
\onecolumn

\section{Derivation of the Empirical SQ Loss}
\label{subsec:SQ_derivation}
The SQ loss can be further expanded as follows:
\begin{align*}
\mathcal{L}_{SQ}&=\frac{1}{2}\iint_{}^{}\left(\hat{p}(y|x)-p(y|x)\right) ^2p(x)dxdy.\\
&=\frac{1}{2}\iint_{}^{}(\hat{p}(y|x))^2p(x)dxdy-\iint_{}^{}\hat{p}(y|x)p(x,y)dxdy+C,
\end{align*}
where $C$ is the constant term. By plugging in $\hat{p}(y|x)=\sum_{a=1}^{M} k_{\sigma}(y, \eta_{a}) f_a(x;\theta)$, the empirical estimate can be written as:
\begin{align*}
\hat{\mathcal{L}}_{SQ}=\frac{1}{2}\sum_i^{}\sum_{a,b}^{}\left(\int_{}^{}k_{\sigma}(\eta_a,y)k_{\sigma}(y,\eta_b)dy\right)f_a(x_i;\theta)f_b(x_i;\theta)-\sum_i^{}\sum_{a}^{}k_{\sigma}(y_i,\eta_a)f_a(x_i;\theta).
\end{align*}
The second term corresponds to the first term in $\hat{\ell}_{SQ}(\sigma)$. For the first term, we have an integral $\int_{}^{}k_{\sigma}(\eta_a,y)k_{\sigma}(y,\eta_b)dy$. With $k_{\sigma}$ having the form of Gaussian density, this can be calculated analytically:
\begin{align*}
\int k(\eta_a, y) k(y, \eta_b) dy &= \left(\frac{1}{2\pi\sigma^2}\right)^{d_y}\int \exp\left(-\frac{1}{2\sigma^2} (|\eta_a - y|^2 + |y - \eta_b|^2)\right) dy\\
&=\left(\frac{1}{2\pi\sigma^2}\right)^{d_y} \int \exp\left(-\frac{1}{4\sigma^2}\left( 4\left|y - \frac{\eta_a+\eta_b}{2}\right|^2+|\eta_a - \eta_b|^2 \right)\right) dy\\
&=\left(\frac{1}{2\pi\sigma^2}\right)^{d_y} \int \exp\left(-\frac{1}{\sigma^2} \left|y - \frac{\eta_a+\eta_b}{2}\right|^2\right) dy\exp\left(-\frac{1}{4\sigma^2} |\eta_a - \eta_b|^2\right)\\
&= \left(\frac{1}{2\pi(\sqrt{2}\sigma)^2}\right)^{d_y}\exp\left(-\frac{|\eta_a - \eta_b|^2}{2(\sqrt{2}\sigma)^2}\right)\\
&=k_{\sqrt{2}\sigma}(\eta_a,\eta_b).
\end{align*}
combining these elements, we obtain:
\begin{align*}
    \hat{\mathcal{L}}_{SQ}=\sum_{i=1}^{n}\left(-2\sum_{a}k_{\sigma}(y_i,\eta_a)f_a(x_i;\theta)+\sum_{a,b}k_{\sqrt{2}\sigma}(\eta_a,\eta_b)f_a(x_i;\theta)f_b(x_i;\theta)\right).
\end{align*}

\section{Proof of \cref{theorem:rkhs_norms}}
\label{subsec:Proof_Kenji}

We first introduce the Fourier expression of RKHS given by a shift invariant integrable kernel $k(x-y)$ on ${\rm I\!R}^d$.  Let $\hat{k}$ denote the Fourier transform of $k(z)$:
\[
\hat{k}(\omega) = \frac{1}{(2\pi)^{d/2} }\int k(z)e^{-\sqrt{-1}
\omega^T z} dz.
\]
It is known \citep[e.g.][]{Girosi1995,Saitoh1997} that a function $g$ on ${\rm I\!R}^d$ is in the corresponding RKHS $\mathcal{H}_k$ if and only if 
\[
\int \frac{|\hat{g}(\omega)|^2}{\hat{k}(\omega)}d\omega < \infty
\]
and its squared RKHS norm is given by 
\[
\| g \|_{\mathcal{H}_k}^2 = \int \frac{|\hat{g}(\omega)|^2}{\hat{k}(\omega)}d\omega.
\]
Note that by Bochner's theorem, the Fourier transform $\hat{k}(\omega)$ takes non-negative values. 

It is well known that, the Fourier transform of Gaussian density kernel $k_\sigma(z) = \frac{1}{(2\pi\sigma^2)^{d/2}}\exp(-\frac{\|z\|^2}{2\sigma^2})$ is given by 
\begin{equation}\label{eq:Gauss_Fourier}
\hat{k}_\sigma(\omega) = \frac{1}{(2\pi)^{d/2}} e^{-\frac{\sigma^2\|\omega\|^2}{2}}.
\end{equation}
Let 
\[
g=\sum_{a=1}^M k_{\sigma}(\cdot,\eta_a)w_a \in\mathcal{H}_\sigma \quad\text{and}\quad 
f=\sum_{a=1}^M k_{\sqrt{2}\sigma}(\cdot,\eta_a)w_a \in\mathcal{H}_{\sqrt{2}\sigma} 
\]
be the two functions in Theorem \ref{theorem:rkhs_norms}. It is easy to see from the general Fourier formulas that
\[
\hat{g}(\omega) = \sum_{a}  w_a e^{-\sqrt{-1}\eta_a^T\omega} \hat{k}_{\sigma}(\omega) 
\]
and thus we obtain from \eqref{eq:Gauss_Fourier}
\[
\| g\|_{\mathcal{H}_{\sigma}}^2 =
 \frac{1}{(2\pi)^{d/2}}  \int  \Bigl| \sum_{a}w_a e^{-\sqrt{-1}\eta_a^T\omega}\Bigr|^2 \exp\Bigl(-\frac{\sigma^2\|\omega\|^2}{2}\Bigr) d\omega.
\]
Similarly, by replacing $\sigma$ with $\sqrt{2}\sigma$, we have 
\[
\| f\|_{\mathcal{H}_{\sqrt{2}\sigma}}^2 =
 \frac{1}{(2\pi)^{d/2}}  \int  \Bigl| \sum_{a}w_a e^{-\sqrt{-1}\eta_a^T\omega}\Bigr|^2 \exp\bigl(-\sigma^2\|\omega\|^2\bigr) d\omega.
\]
It follows from $\exp\bigl(-\sigma^2\|\omega\|^2\bigr)\leq \exp\Bigl(-\frac{\sigma^2\|\omega\|^2}{2}\Bigr)$ for any $\omega$ that 
\[
\| f\|_{\mathcal{H}_{\sqrt{2}\sigma}}^2 \leq \| g\|_{\mathcal{H}_{\sigma}}^2,
\]
which completes the proof.

\section{Details on Experiments in \cref{exp:toy}}\label{toy_details}
\subsection{Data Generating Process}
The visualization of the toy datasets used in the experiment is shown in \cref{Toydata}. The generating process is provided in the following:
\begin{figure*}[ht]
\vskip 0.2in
\begin{center}
\centerline{\includegraphics[width=1.0\textwidth]{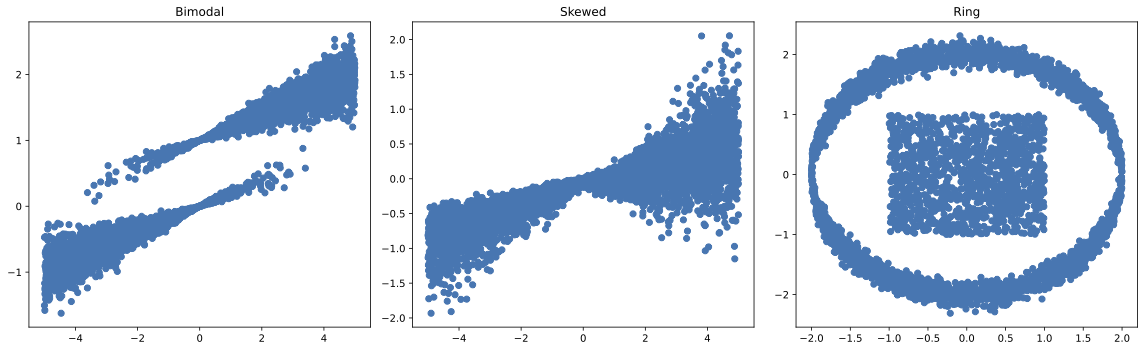}}
\caption{Visualization of toy datasets}
\label{Toydata}
\end{center}
\vskip -0.2in
\end{figure*}

{\bf Bimodal:} 

$y=0.2x+P+\epsilon$,
where $x\sim U(-5,5)$, $P\sim Binomial\left(\frac{1}{1+\exp(-1.5x)}\right)$, and $\epsilon\sim N(0, (0.05x)^2)$.

{\bf Skewed:} 

$y\sim SkewNormal(\xi(x),\omega(x),\alpha(x))$, 
where $x\sim U(-5,5)$, $\xi(x)=0.1x$, $\omega(x)=0.1|x|+0.05$, and $\alpha(x)=-8+8\cdot\left(\frac{1}{1+\exp(-x)}\right)$. Here, $\xi(x)$ corresponds to location, $\omega(x)$ to scale, and $\alpha(x)$ to skewness.

{\bf Ring:} 
\begin{align*}
y=\begin{cases}\begin{cases}U(-1, 1)   & P_1 = 1/2\\Ring(x) & P_1 = 1/2\end{cases} & |x| \leq1 \\Ring(x) & |x| > 1\end{cases},
\end{align*}
and 
\begin{align*}
Ring(x)=\begin{cases}2 \cdot \sin(\arccos(x/2))+\epsilon    & P_2 = 1/2\\2 \cdot \sin(-\arccos(x/2))+\epsilon & P_2 = 1/2\end{cases},
\end{align*}
where $x\sim U(-2,2)$ and $\epsilon\sim N(0,0.1^2)$.

\subsection{Architecture and Hyperparameter Choices}
We used NNs with two fully-connected hidden layers, each containing 50 ReLU activation units. For the optimizer, we used AdamW \citep{loshchilov2018decoupled}. Other architectural and hyperparameter choices for each model are provided below:

{\bf Proposals:} 
We set the number of location points $M=100$, and $\eta_a$ were chosen as uniformly spaced grid points within the closed interval bounded by the minimum and maximum values observed in the training data. The learning rate was set to 1e-4, the batch size was set to 50, and the number of training epochs was set to 1000, and $\sigma$ was initialize to 1.0. For Proposals, we have a final layer that outputs the weights on the location points. 

{\bf DFs:} 
The regularization parameter $\lambda$ was set to 0.1, the learning rate was set to 1e-4, the batch size was set to 50, and the number of training epochs was set to 1000. For the kernel, we used Gaussian kernel. For DFs, the second hidden layer also corresponds to the final layer that outputs the features. 

{\bf MDN:} 
The number of mixing component was set to 10, the learning rate was set to 1e-4, the batch size was set to 50, and the number of training epochs was set to 1000. For MDN, we have three final layers that output means, variances and mixing weights. 

{\bf NF:} 
We used the conditional version of the autoregressive Neural Spline Flow \cite{durkan2019neural} implementation by \citet{Stimper2023}. The number of flows were set to 5, the learning rate was set to 1e-3, the batch size was set to 256,  and the number of training epochs was set to 100. For NF, the number of NNs mirrors the number of flows, resulting in 5 NNs in this case.

{\bf CARD:} 
We used the implementation provided by \citet{han2022card}, and a GitHub repository \url{https://github.com/lightning-uq-box/lightning-uq-box}. For the conditional mean estimator, we applied the same NN architecture as other methods, and the learning rate was set to 1e-3, the batch size was set to 256, and the number of training epochs was set to 100. For denoising diffusion model, we used the same architecture used in the toy data experiments in \citet{han2022card}: NN based on three fully-connected hidden layers with 128 units. The learning rate was set to 1e-3, the batch size was set to 256, the number of time steps was set to 1000, and the number of training epochs was set to 5000. Note that for CARD, we first optimize the conditional mean estimator, and then use this to guide the next optimization procedure for the denoising diffusion model.

\section{Details on Experiments in \cref{exp:uci}}\label{uci_details}
\subsection{UCI Datasets}
We report the number of data points and input features for each dataset in \cref{datainfo}. We excluded the Yacht dataset due to its limited size of only 308 data points, and the Wine dataset because its output variable consists of only 5 discrete values and exhibits near-linear relationships.
\begin{table}[ht]
\caption{\label{datainfo}(number of data points , number of input features) for each dataset}\vspace{-2.5mm}
\begin{center}
\resizebox{\textwidth}{!}{
\begin{tabular}{@{}ccccccccccc@{}}
\toprule[1.5pt]
Dataset  & Boston      & Concrete     & Energy    & Kin8nm       & Naval          & Power        & Protein       & Year            \\ \midrule
$ $ & $(506, 13)$ & $(1030, 8)$ & $(768, 8)$ & $(8192, 8)$ & $(11934, 16)$ & $(9568, 4)$ & $(45730, 9)$ &$(515345, 90)$     \\ \bottomrule[1.5pt]
\end{tabular}}
\end{center}
\end{table}

\subsection{Architecture and Hyperparameter Choices}
For Boston, Concrete, Energy and Kin8nm, we used NNs with three fully-connected hidden layers, each containing 50 ReLU activation units. For Naval, Power, Protein and Year, we used NNs with three fully-connected hidden layers, each containing 100 ReLU activation units. For the optimizer, we used AdamW \citep{loshchilov2018decoupled}. We report the learning rate, the batch size and the number of training epochs in \cref{tab:uci_learning_rate}, \cref{tab:uci_batch_size}, and \cref{tab:uci_epochs}, respectively. Other architectural and hyperparameter choices for each model are provided below:

{\bf Proposals:} 
We used Spectral Normalization (SN) \citep{miyato2018spectral} in the second hidden layer and the final output layer. We set the number of location points $M=100$, and $\eta_a$ were chosen as uniformly spaced grid points within the closed interval bounded by the minimum and maximum values observed in the training data. The bandwidth $\sigma$ was initialized to 1.0. For Year dataset, when iteratively optimizing kernel parameter (Proposal-Iterative), we opted to update $\sigma$ every 4 steps.

{\bf DFs:} 
We used SN in the third hidden layer (which also corresponds to the final output layer). The regularization parameter $\lambda$ was set to 0.1, and for the kernel we used Gaussian kernel.

{\bf MDN:} 
We used SN in the third hidden layer. The number of mixing components was set to 10, except for Protein and Year where it was set to 5.

{\bf NF:} 
We used the conditional version of the autoregressive Neural Spline Flow \cite{durkan2019neural} implementation by \citet{Stimper2023}. For NF, we found that the model can easily overfit, and so we used slightly different NN architectures from the others: For Boston, Concrete, Energy and Kin8nm, we used NNs with \textbf{two} fully-connected hidden layers, each containing 50 ReLU activation units. For Naval, Power, Protein and Year, we used NNs with three fully-connected hidden layers, each containing \textbf{50} ReLU activation units. The number of flows were set to 5, and the number of training epoch was set as 10 to mitigate overfitting.

{\bf CARD:} We directly adopt the results presented in \citet{han2022card}.

\begin{table}[ht]
\caption{\label{tab:uci_learning_rate}Learning rates}\vspace{1mm}
\begin{center}
\resizebox{6cm}{!}{
\begin{tabular}{@{}l|ccccc@{}}
\toprule[1.5pt]
         & Proposals           & DFs       & MDN       &NF   \\ \midrule
Boston   & $0.0005$       & $0.0005$   & $0.0005$   & $0.001$       \\
Concrete & $0.0005$       & $0.0005$   & $0.0005$   & $0.001$          \\
Energy   & $0.0005$       & $0.0005$   & $0.0005$   & $0.001$         \\
Kin8nm   & $0.0005$       & $0.0005$   & $0.0005$   & $0.001$          \\
Naval    & $0.001$       & $0.001$   & $0.001$   & $0.001$         \\
Power    & $0.001$       & $0.001$   & $0.001$   & $0.001$         \\
Protein  & $0.001$      & $0.001$  & $0.001$  & $0.001$       \\
Year     & $0.001$      & $0.001$  & $0.001$  & $0.001$  \\ \bottomrule[1.5pt]
\end{tabular}
}
\end{center}
\end{table}

\begin{table}[ht]
\caption{\label{tab:uci_batch_size}Batch sizes}\vspace{1mm}
\begin{center}
\resizebox{6cm}{!}{
\begin{tabular}{@{}l|ccccc@{}}
\toprule[1.5pt]
         & Proposals  & DFs     & MDN    &NF   \\ \midrule
Boston   & $32$       & $32$   & $32$   & $50$       \\
Concrete & $32$       & $32$   & $32$   & $50$          \\
Energy   & $32$       & $32$   & $32$   & $50$         \\
Kin8nm   & $100$       & $100$   & $100$   & $100$          \\
Naval    & $256$       & $100$   & $100$   & $256$         \\
Power    & $100$       & $100$   & $100$   & $256$         \\
Protein  & $256$      & $100$  & $256$  & $256$       \\
Year     & $512$      & $256$  & $512$  & $256$  \\ \bottomrule[1.5pt]
\end{tabular}
}
\end{center}
\end{table}

\begin{table}[ht]
\caption{\label{tab:uci_epochs}Training epochs}\vspace{1mm}
\begin{center}
\resizebox{6cm}{!}{
\begin{tabular}{@{}l|ccccc@{}}
\toprule[1.5pt]
         & Proposals  & DFs       & MDN    &NF   \\ \midrule
Boston   & $500$       & $500$   & $250$   & $10$       \\
Concrete & $500$       & $500$   & $250$   & $10$          \\
Energy   & $500$       & $500$   & $250$   & $10$         \\
Kin8nm   & $500$       & $500$   & $250$   & $10$          \\
Naval    & $500$       & $500$   & $250$   & $10$         \\
Power    & $500$       & $500$   & $250$   & $10$         \\
Protein  & $500$      & $500$  & $250$  & $10$       \\
Year     & $50$      & $50$  & $50$  & $10$  \\ \bottomrule[1.5pt]
\end{tabular}
}
\end{center}
\end{table}

\subsection{Additional Results: RMSE}
\cref{tab:RMSE} presents the RMSE values for each dataset. RMSE was computed using the same samples employed for QICE metric evaluation. Results show that CARD outperforms other methods in most cases. This aligns with expectations, as the initial step of CARD involves training a conditional mean estimator. Notably, other methods do not include explicit conditional mean estimation during training or within their objective functions. Consequently, we interpret comparable RMSE values to CARD as evidence that the generated samples avoids any pathological behaviors that may potentially exploit the QICE metric. 
\begin{table}
\caption{\label{tab:RMSE}RMSE for UCI datasets. For Kin8nm and Naval dataset, values are multiplied by $100$.}\vspace{1mm}
\begin{center}
\resizebox{13cm}{!}{
\begin{tabular}{@{}l|ccccccc@{}}
\toprule[1.5pt]
Dataset  &                     &                     & RMSE $\downarrow$                &                     &                     \\
         & Proposal-Iterative                 & Proposal-Joint          & DF-med  & DF-0.1      & MDN     &NF               & CARD              \\ \midrule
Boston   & $3.56\pm 1.02$  & $3.45\pm 1.08$  & $3.40\pm 0.92$ & $4.01\pm 1.08$  & $3.36\pm 1.14$  & $4.14\pm 1.18$ & $\bm{2.61\pm 0.63}$  \\
Concrete & $6.40\pm 0.83$  & $5.98\pm 0.63$  & $5.84\pm 0.54$ & $7.74\pm 0.76$  & $5.62\pm 0.58$  & $7.22\pm 0.62$ & $\bm{4.77\pm 0.46}$  \\
Energy   & $1.19\pm 0.21$ & $0.99\pm 0.16$  & $0.69\pm 0.13$ & $2.67\pm 0.25$  & $2.24\pm 0.43$   & $2.88\pm 0.30$ & $\bm{0.52\pm 0.07}$  \\
Kin8nm   & $8.19\pm 0.23$ & $8.11\pm 0.29$ & $7.11\pm 0.21$ & $9.52\pm 0.27$ & $6.95\pm 0.22$  & $7.98\pm 0.32$ & $\bm{6.32\pm 0.18}$ \\
Naval    & $0.09\pm 0.02$  & $0.17\pm 0.03$ & $\bm{0.01\pm 0.00}$ & $0.21\pm 0.02$ & $0.08\pm 0.05$  & $0.36\pm 0.08$ & $0.02\pm 0.00$ \\
Power    & $3.96\pm 0.20$  & $3.89\pm 0.19$ & $4.08\pm 0.16$ & $4.56\pm 0.17$  & $\bm{3.79\pm 0.17}$   & $4.19\pm 0.18$ & $3.93\pm 0.17$  \\
Protein  & $3.97\pm 0.05$  & $3.93\pm 0.03$  & $4.77\pm 0.05$ & $4.83\pm 0.04$  & $3.79\pm 0.04$  & $4.38\pm 0.04$  & $\bm{3.73\pm 0.01}$  \\
Year     & $8.82\pm$ NA       & $8.82\pm$ NA        & $8.88\pm$ NA     & $8.92\pm$ NA       & $8.79\pm$ NA       & $8.80\pm$ NA  & $\bm{8.70\pm}$ NA         \\ \bottomrule[1.5pt]
\end{tabular}
}
\end{center}
\end{table}

\section{Details on RL Experiments}
\subsection{Environmnets}
We briefly describe environments used in our experiments:

{\bf CartPole-v1:} An agent controls a cart on a frictionless track, aiming to balance a pole. The state space is four-dimensional, encompassing cart position, velocity, pole angle, and pole tip velocity. The agent can choose to push the cart left or right, receiving a $+1$ reward per balanced time step. Episodes terminate when the pole angle exceeds $\pm12$, the cart reaches the track edge, or the episode exceeds 500 steps.

{\bf Acrobot-v1:} This environment features a two-linked pendulum with a controllable joint. The agent aims to swing the outer link's end to a specific height. The six-dimensional state describes joint angles and velocities, and the agent can apply no torque, torque left, or torque right. It receives a -1 reward per step before reaching the goal, and episodes end upon reaching the goal or exceeding 500 steps.

{\bf MountainCar-v0:} This environment challenges an agent to drive an under-powered car up the mountain to the right. The agent must gain momentum by moving back and forth to achieve this goal. The state comprises the car's position and velocity, and the agent can push left, do nothing, or push right. Each step incurs a -1 reward until reaching the goal position, and episodes end upon reaching the goal or exceeding 200 steps. 
\subsection{Architecture and Hyperparameter Choices}
We first describe architecture and hyperparameter choices shared across all methods: The NN architecture comprised two fully-connected hidden layers, each containing 50 ReLU activation units. For the optimizer, we used Adam \citep{kingma2015adam}, and learning rates were adjusted for each environment: 1e-4 for CartPole and 1e-3 for Acrobot and MountainCar. A discount factor $\gamma$ of 0.99 and batch size of 32 were used consistently. The replay buffer held 10,000 experiences, with parameter updates occurring every 2 steps and target network updates every 100 steps (except where noted). For exploration, an $\epsilon$-greedy policy with linear decay was implemented, starting $\epsilon$ with 1.0 and decaying to 0.01 over 10,000 steps. Finally, agents interact with the environment for a total of 500,000 steps, and were evaluated on an independent test episode every 100 steps with $\epsilon=0.001$. Also note that to isolate the influence of our method's design choices, we refrained from employing common RL techniques such as double Q-learning \citep{vanhasselt2015doubledqn} and prioritized experience replay \citep{schaul2015prioritized}. Other architectural and hyperparameter choices for each model are provided below:

{\bf Proposal:} We set $\eta_a$ as uniform grid of 51 points, ranging from -100 to 100. Consequently, the final output layer dimension of NN becomes $51\times |A|$ where $|A|$ represents the action space size. We also apply softmax fuction to normalize these 51 points associate with each action, which we find to result in improved performance. For kernel, we used Gaussian kernel. Following \citet{biggs2023mmdfuse}, the distribution over kernels $\omega$ was defined as a uniform distribution over collections of Gaussian kernels with 10 different bandwidths $\sigma>0$. These bandwidths were selected from a uniform grid spanning the interval between half the 5th percentile and half the 95th percentile of the distance values $\left\{\|\eta_a-\eta^{\prime}_a\|\right\}$.

{\bf CDQN:} We set $\delta_a$ as uniform grid of 51 atoms, ranging from -100 to 100. Consequently, the final output layer dimension of NN becomes $51\times |A|$. Softmax fuction is applied to normalize these 51 atoms associate with each action. For CartPole, we set the target update period to 1000, as this led to improved performance.

{\bf MMDQN:} We set the number of particles learned by NN to 51.  Consequently, the final output layer dimension of NN becomes $51\times |A|$. For CartPole and MountainCar, we set the target update period to 1000, and also scale the reward by 0.1 (aiding particle learning by NN), as this led to improved performance.  For kernel, we followed \citet{Nguyen-Tang_Gupta_Venkatesh_2021} and used (uniform) mixture of 10 Gaussian kernels, with bandwidths set as uniformly spaced values between 1 and 100. When the reward was scaled by 0.1, this bandwidth range was adjusted to 1 to 10.

{\bf DQN:} Final output layer dimension of the NN is $|A|$.

\subsection{Additional Results on Proposed Model}
\begin{figure*}[ht]
\vskip 0.2in
\begin{center}
\centerline{\includegraphics[width=1.0\textwidth]{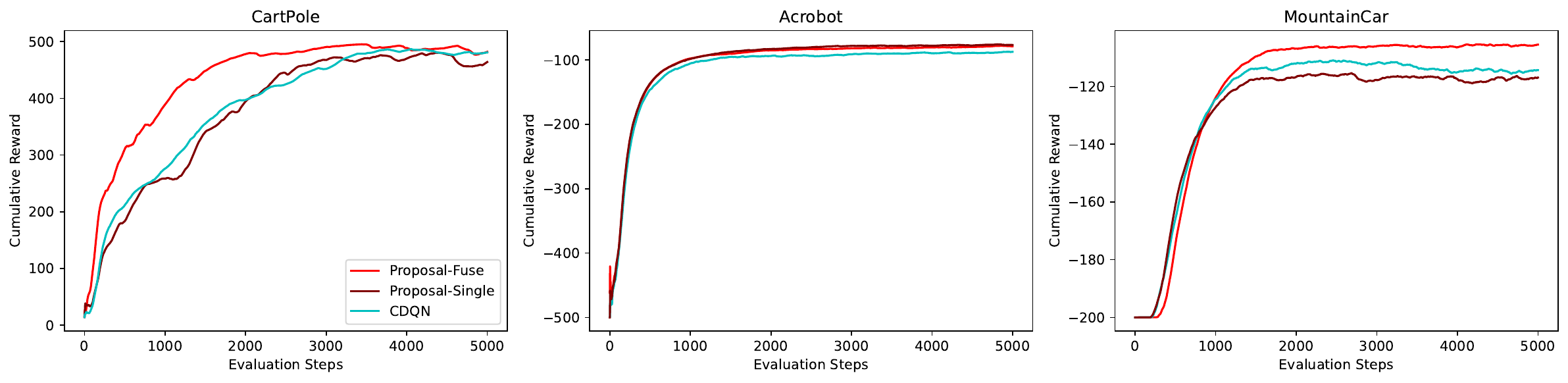}}
\caption{Comparison of our proposed methods using single and fused kernels. We report the mean of cumulative rewards across 10 independent runs.}
\label{proposal_comparison}
\end{center}
\vskip -0.2in
\end{figure*}

To investigate the impact of our fusing strategy (Proposal-Fuse), we conducted a comparative analysis with a variant of our proposal that employs a single bandwidth parameter (Proposal-Single). We used the same architecture and hyperparameter as CDQN, and we set the bandwidth to 10. The result shown in \cref{proposal_comparison} demonstrates a significant performance improvement when using the fusing strategy. Without the fusing, Proposal-Single is slightly worse than CDQN on CartPole and MountainCar. This highlights both the efficacy of the fusing approach and the critical role of kernel hyperparameter selection in DRL contexts.

\end{document}